\documentclass[11pt]{article}

\usepackage{geometry}
\usepackage{amsmath, amsfonts, bm, amssymb, enumitem, tabularx}
\usepackage{amsfonts, graphicx, grffile,fullpage, float, latexsym, mathtools}
\usepackage{hyperref}
\hypersetup{
    colorlinks=true, 
    citecolor=black,
     linkcolor=blue,
     urlcolor=blue,
    linktoc=all,     
    linkcolor=black,  
}
\usepackage{cleveref}
\usepackage{multicol}
\usepackage{tikz}

\usepackage[sort, numbers, compress]{natbib}

\usepackage{Shorthands}

\title{A Data-Driven Approach to Synthesizing Dynamics-Aware Trajectories for Underactuated Robotic Systems}
\author{Anusha Srikanthan$^{1}$, Fengjun Yang$^{1}$, Igor Spasojevic$^{1}$, \\ Dinesh Thakur$^{2}$, Vijay Kumar$^{1}$, Nikolai Matni$^{1}$ 
\thanks{This research is in part supported by NSF award CPS-2038873, NSF CAREER award ECCS-2045834, NSF Grant CCR-2112665, and a Google Research Scholar award.
$^{1}$are with GRASP Lab, University of Pennsylvania, Philadelphia, USA
        {\tt\small \{sanusha, fengjun, igorspas, kumar, nmatni\}@seas.upenn.edu}.
$^{2}$was with GRASP Lab at the time of writing this paper.}}

\begin{document}

\maketitle

\begin{abstract}
We consider joint trajectory generation and tracking control for under-actuated robotic systems.
A common solution is to use a layered control architecture, where the top layer uses a simplified model of system dynamics for trajectory generation, and the low layer ensures approximate tracking of this trajectory via feedback control.
While such layered control architectures are standard and work well in practice, selecting the simplified model used for trajectory generation typically relies on engineering intuition and experience. 
In this paper, we propose an alternative data-driven approach to \emph{dynamics-aware trajectory generation}.
We show that a suitable augmented Lagrangian reformulation of a global nonlinear optimal control problem results in a layered decomposition of the overall problem into trajectory planning and feedback control layers.
Crucially, the resulting trajectory optimization is dynamics-aware, in that, it is modified with a \emph{tracking penalty regularizer} encoding the dynamic feasibility of the generated trajectory.
We show that this tracking penalty regularizer can be learned from system rollouts for independently-designed low layer feedback control policies, and instantiate our framework in the context of a unicycle and a quadrotor control problem in simulation.
Further, we show that our approach handles the sim-to-real gap through experiments on the quadrotor hardware platform without any additional training. For both the synthetic unicycle example and the quadrotor system, our framework shows significant improvements in both computation time and dynamic feasibility in simulation and hardware experiments.
\end{abstract}

\section{Introduction}

Modularity is a guiding principle behind the design of numerous autonomous platforms. 
For example, the autonomy stack of a typical robot consists of separate modules for perception, planning, and control~\cite{paden2016survey}.
In spite of the requirement of safely executing tasks in real time with limited on board computational resources, such modules usually operate at different frequencies and levels of abstraction. 
Roughly speaking, higher levels of abstraction allow for faster decision making.  
However, if the degree of abstraction varies among the different modules beyond a suitable threshold, the system as a whole can behave in unexpected, unsafe ways. 
By and large, choosing the right level of abstraction in robotics applications has remained somewhat of an art. 
We focus on developing a quantitative method of bridging the potential mismatch between the trajectory planning and control modules in a data-driven manner.

Although trajectory planning and control have been among the most extensively studied areas of robotics, numerous problems remain to be solved. 
In particular, graph-search-based path planning algorithms can find it challenging to account for complex nonlinear system dynamics. Similarly, real-time optimization-based methods for generating trajectories typically use a simplified or a reduced order dynamics model of the agent.  
In contrast, low-level feedback control policies often rely on more accurate, detailed dynamics of the system being controlled in order to track a reference trajectory planned by some of the aforementioned approaches.
While intuitive and conceptually appealing, this layered approach only works well if the outputs of higher layers are compatible with the abilities of lower layers.  

In this paper, we focus on the interplay between trajectory generation and feedback control.  
Rather than imposing such a layered architecture on the control stack, we show that it can be \emph{derived} via a suitable relaxation of a global nonlinear optimal control problem that jointly encodes both the trajectory generation and feedback control problems. 
Crucially, the resulting trajectory generation optimization problem is dynamics-aware, in that it is modified with a \emph{tracking penalty regularizer} that encodes the dynamic feasibility of a generated trajectory.  
While this tracking penalty does not in general admit a closed-form expression, 
we show that it can be interpreted as a cost-to-go. Hence, it can be learned from system roll-outs for any feedback control policy by leveraging tools from the learning literature. Finally we evaluate our framework using unicycle and quadrotor control, and compare our approach in simulation to standard approaches to quadrotor trajectory generation. 
Our extensive experiments demonstrate that our data-driven dynamics-aware framework allows for faster computation of trajectories that can be tracked accurately in both simulation and hardware. 
Our contributions are as follows:
\begin{itemize}
    \item We derive a layered control architecture composed of a dynamics-aware trajectory generator top layer, and a feedback control low layer.  In contrast to existing work, our trajectory generation problem is naturally dynamics-aware, and includes a tracking penalty regularizer that encodes the ability of the low-layer feedback control policy to track a given reference trajectory.
    \item We show how this tracking penalty can viewed as the cost-to-go for a particular system, and hence be learned from system rollouts. 
    \item We apply our data-driven dynamics-aware trajectory generation framework to both a unicycle and a quadrotor control problem. We demonstrate that our approach generates aggressive and easy to track trajectories compared to standard methods for the two systems in consideration. 
\end{itemize}

In what follows, we first formulate the dynamics-aware trajectory planning problem in Section \ref{sec:problem-formulation} and introduce the unicycle and waypoint tracking problem as running examples. In Section \ref{sec:layeredarch}, we introduce a result that shows how a relaxation of the underlying nonlinear controls problem naturally leads to a trajectory optimization problem that includes a regularizer that captures the tracking cost of the given controller. 
We also describe our supervised learning approach to learn the cost-to-go function that characterizes the feedback control layer's ability to track a given reference trajectory. In Section \ref{sec:method-uni}, we apply our approach to dynamics-aware trajectory generation to both the unicycle and quadrotor systems. We present two compelling simulation experiments in Section \ref{sec:experiments} to show that our method leverages previous trajectory data to approximate the cost-to-go function and the learned function can be applied to generate easy-to-track trajectories before discussing the results and future work in Section \ref{sec:conclusion}.

\section{Related Work}
Our work builds on the literature of multi-rate/hierarchical control, data-driven nonlinear model-predictive-control (NMPC), and trajectory generation for quadrotors.  Here we attempt to provide an overview of most directly relevant related works from these different branches of the robotics literature.

\subsection{Multi-rate and hierarchical control}
There is a rich literature on combining trajectory generation with low-level control, see for example~\cite{rosolia2020multi, herbert2017fastrack, wabersich2018linear, yin2020optimization, singh2018robust, singh2017robust, gurriet2018towards} and references therein.  While these results offer differing degrees of guarantees and generality, we note that none of them derive the layered control architecture that they propose.  Rather, modification to either the trajectory generation or tracking problems are made to ensure that the chosen interplay between the two layers leads to desirable results.  In contrast, we derive this layered structure, and show how this naturally leads to the inclusion of a tracking penalty regularizer in the trajectory generation problem.  Our work is complementary to the existing literature due to the fact that modifying any of the proposed trajectory generation optimization problems in~\cite{rosolia2020multi, herbert2017fastrack, wabersich2018linear, yin2020optimization, singh2018robust, singh2017robust, gurriet2018towards}  with our proposed regularizer will only lead to more dynamically feasible trajectories.

Another closely-related line of work in this spirit is the ``Layering as Optimization Decomposition'' framework proposed in~\cite{chiang2007layering}, which shows that network utility maximization problems can be suitably relaxed to recover the layered architecture of network control protocol stacks.  This approach was extended to linear optimal control problems in~\cite{matni2016theory}, 
but was limited to LQR control for which the tracking penalty admits a closed-form expression.  In contrast, we significantly generalize these results to nonlinear dynamical systems and present a data-driven approach to approximating the tracking penalty.

\subsection{Data-driven NMPC}
Due to the underlying difficulty of solving a NMPC problem that jointly encodes trajectory generation and low-level control, data-driven approaches to improve controller performance on tracking tasks have emerged. Broadly, learning can be applied to: (i) directly obtain the control policy \cite{levine2014learning, levine2016end}, (ii) learn uncertainties in the dynamics and the cost function used for NMPC \cite{kabzan2019learning, williams2017information, ostafew2016robust, rosolia2019learning}, and (iii) learn a low dimensional state representation for NMPC from high dimensional data \cite{kaufmann2019beauty, drews2017aggressive, song2022policy}. Our work broadly fits into this overall line of work in that we propose a data-driven method to learn a tracking penalty regularizer that directly encodes the closed-loop dynamics' ability to track a generated trajectory using offline trajectories.  To the best of our knowledge, ours is the first approach to suggest this compact encoding of the low layer closed-loop dynamics into a cost-to-go function. 

\subsection{Trajectory generation for quadrotors}

In general, trajectory generation for quadrotors is a computationally challenging problem.
%
The landmark paper \cite{mellinger2011minimum} established \textit{differential flatness} \cite{fliess1995flatness} of quadrotor dynamics. 
It showed that trajectories of position and yaw angle of the quadrotor, i.e., the flat outputs, may be specified \textit{independently} of one another, and that their time derivatives yield the underlying trajectory of states and inputs required to induce them. 
Additionally, \cite{mellinger2011minimum} initiated a line of work~\cite{RichterBryRoy2016,HehnDAndreaRealTimeTRO15,MuellerDAndreaCompEffPrimitiveTRO15,LiuSE3SearchRAL18} using piecewise polynomials to represent trajectories of flat outputs.
Nevertheless, these approaches decouple trajectory generation and tracking control. 
As a consequence, there is no model of the specific hardware used for control introduced in the planning layer. 
Our current work, on the other hand, provides a principled way of generating trajectories cognizant of the dynamic capabilities of the closed-loop robotic system.

\section{Problem Formulation} \label{sec:problem-formulation}
 
We consider a finite-horizon, discrete-time nonlinear dynamical system
\begin{equation} \label{eq:dynamics}
x_{t+1} = f(x_t, u_t), \ t=0,\dots, N,
\end{equation}
with state $x_t \in \mathcal{X} \subseteq \mathbb{R}^n$ and control input $u_t \in \mathcal{U} \subseteq \mathbb{R}^k$ at time $t$. 
Our task is to solve the following constrained optimal control problem (OCP):
\begin{equation} \label{prob:master-problem}
\begin{array}{rl}
    \underset{x_{0:N},u_{0:N-1}}{\mathrm{minimize}} & \mathcal{C}(x_{0:N}) + \sum_{t=0}^{N-1}\|D_t u_t\|_2^2  \\
     \text{s.t.} & x_{t+1} = f(x_t,u_t), \, t=0,\dots,N,\\
     & x_{0:N} \in \mathcal{R},
\end{array}    
\end{equation}
where $\mathcal{C} : \mathcal{X}^{N+1} \rightarrow \mathbb{R}$ is a trajectory cost function, $D_0, D_1, ...., D_{N-1} \in \mathbb{R}^{l \times k}$
are matrices that penalize control effort, and $\mathcal{R}$ defines the feaible region of $x_{0:N}$. %

OCPs of the form \eqref{prob:master-problem} are 
an essential component of MPC schemes for robotic applications. In such settings, the trajectory cost function $\mathcal{C}$ is typically chosen to e.g., capture high-level task objectives or reward smooth trajectories, whereas the state constraint $\mathcal{R}$ is often used to encode e.g., obstacle avoidance, waypoint constraints, or other mission-specific requirements. %
In the generality stated above, the OCP~\eqref{prob:master-problem} is 
difficult to solve exactly except in the simplest of cases.  Under suitable regularity assumptions, good heuristics exist for finding an approximate solution. However, due to their computational complexity, these heuristics typically lead to the solve time being unacceptably large for applications with fast dynamics such as quadrotor control.

A number of works in the robotics literature approach the computational complexity by using a \emph{layered control architecture} to decompose OCP~\eqref{prob:master-problem} into tractable subproblems.  
For example, a two-layer approach would solve a reference trajectory generation problem at the \emph{top planning layer} using simplified dynamics (typically at a slower frequency). This reference trajectory would then be sent to the \emph{low tracking layer} where a feedback control policy, operating in real time, attempts to follow the reference trajectory.

While conceptually appealing, the above approach has several shortcomings. The critical one is the lack of guarantees that the generated reference trajectories can be adequately tracked by the feedback control policy. This could be due to unmodelled dynamics, saturation limits, etc., of the hardware in use for control. In this paper, we address this shortcoming by \emph{deriving} a layered architecture via a relaxation of the original OCP~\eqref{prob:master-problem}, that naturally leads to a \emph{closed-loop dynamics-aware} trajectory generation problem.

\section{Layering as Optimal Control Decomposition} \label{sec:layeredarch}

We show how a suitable relaxation of the OCP~\eqref{prob:master-problem} naturally results in a layered control architecture.  Such an optimization decomposition approach to layered control was first introduced in \cite{matni2016theory} for linear-quadratic control. In this section, we extend it to general nonlinear systems. 

\subsection{An augmented Lagrangian relaxation}
We first introduce a redundant reference trajectory variable $r_{0:N}$ constrained to equal the state trajectory, i.e., satisfying $x_{0:N}=r_{0:N}$, to the OCP~\eqref{prob:master-problem}:
\begin{equation}
    \begin{aligned}
        \underset{r_{0:N}, x_{0:N}, u_{0:N-1}}{\mathrm{minimize}}&\quad \mathcal{C}(r_{0:N}) + \sum_{t=0}^{N-1}\lVert D_t u_{t} \rVert_2^2 \\
        \text{s.t.}&\quad x_{t+1} = f(x_t, u_t), \\
        & \quad r_{0:N} \in \mathcal{R}, \\
        & \quad x_{0:N} = r_{0:N}.
    \end{aligned}
\end{equation}
We then relax this redundant equality constraint to a soft-constraint in the objective function, resulting in the augmented Lagrangian reformulation:
\begin{equation} \label{prob:relaxed-problem}
    \begin{array}{rl}
        \underset{r_{0:N}}{\mathrm{min.}}& \mathcal{C}(r_{0:N}) + \underset{x_{0:N}, u_{0:N-1}}{\mathrm{min.}} \sum_{t=0}^{N-1}\left(\lVert D_t u_{t} \rVert_2^2 + \rho \lVert r_t - x_t \rVert_2^2\right) + \rho \lVert r_N - x_N \rVert_2^2\\
        \text{s.t.}&r_{0:N} \in \mathcal{R}, \quad \quad \quad \text{ s.t. } x_{t+1} = f(x_t, u_t)
    \end{array}
\end{equation}
where the weight $\rho>0$ specifies the soft-penalty associated with the constraint $r_{0:N}=x_{0:N}$. Furthermore, we have strategically grouped terms to highlight the nested structure of the resulting optimization problem. Immediately, problem~\eqref{prob:relaxed-problem} admits a layered interpretation: the inner minimization over state and input trajectories $x_{0:N}$ and $u_{0:N-1}$ is a traditional feedback control problem, seeking to optimally track the reference trajectory $r_{0:N}$. The outer optimization over the trajectory $r_{0:N}$ seeks to optimally ``plan'' a reference trajectory for the inner minimization to follow. 
To further highlight the layered nature of the resulting relaxation, we define the tracking penalty
\begin{equation}\label{eq:tracking-cost}
\begin{aligned}
    g_{\rho}^{track}(x_0, r_{0:N}) :=
    \underset{x_{0:N}, u_{0:N-1}}{\mathrm{min}}&\quad \sum_{t=0}^{N-1}\left(\lVert D_t u_{t} \rVert_2^2 + \rho \lVert r_t - x_t \rVert_2^2\right) + \rho \lVert r_N - x_N \rVert_2^2\\
        \text{s.t.}&\quad \text{dynamics \eqref{eq:dynamics}}. 
\end{aligned}
\end{equation}
The tracking penalty $g_{\rho}^{track}(x_0, r_{0:N})$ captures how well a given trajectory $r_{0:N}$ can be tracked by a low layer control sequence $u_{0:N-1}$ given the initial condition $x_0$, and is naturally interpreted as the cost-to-go associated with an augmented system (see \S\ref{sec:learning}).  We observe that the optimal control problem defining the tracking cost~\eqref{eq:tracking-cost} is a standard nonlinear reference tracking problem with quadratic cost, and can be approximately solved using tools from nonlinear feedback control~\cite{roulet2019iterative}.  We therefore let $\pi(x_t,r_{0:N})$ denote the feedback control policy which (approximately) solves problem~\eqref{eq:tracking-cost}, and use $g_{\rho,\pi}^{track}(x_0, r_{0:N})$ to denote the resulting cost-to-go that it induces. While a closed-form expression for the tracking penalty $g_{\rho,\pi}^{track}(x_0, r_{0:N})$ is only available in special cases, e.g., see~\cite{matni2016theory} for the linear quadratic control case, we show in \S\ref{sec:learning} that it can be learned from data.

Assuming that an accurate estimate of the tracking penalty can be obtained, the OCP~\eqref{prob:relaxed-problem} can now be reduced to the \emph{static} optimization problem (i.e., without any constraints enforcing the dynamics~\eqref{eq:dynamics}):
\begin{equation} \label{prob:layered-problem}
    \begin{aligned}
        \underset{r_{0:N}}{\mathrm{minimize}}&\quad \mathcal{C}(r_{0:N}) + g_{\rho, \pi}^{track}(x_0, r_{0:N}) \\
        \text{s.t.} & \quad r_{0:N} \in \mathcal{R}.
    \end{aligned}
\end{equation}
We may view~\eqref{prob:layered-problem} as a family of trajectory optimization problems parametrized by $\rho$. In the limit as $\rho \nearrow \infty$, optimal trajectories prioritize the reference tracking performance. On the other hand, in the limit as $\rho \searrow 0$, the optimal trajectories minimize the $\mathcal{C}$ cost oblivious to the dynamics constraints.

\begin{figure}[t]
    \centering
    \includegraphics[width=\columnwidth]{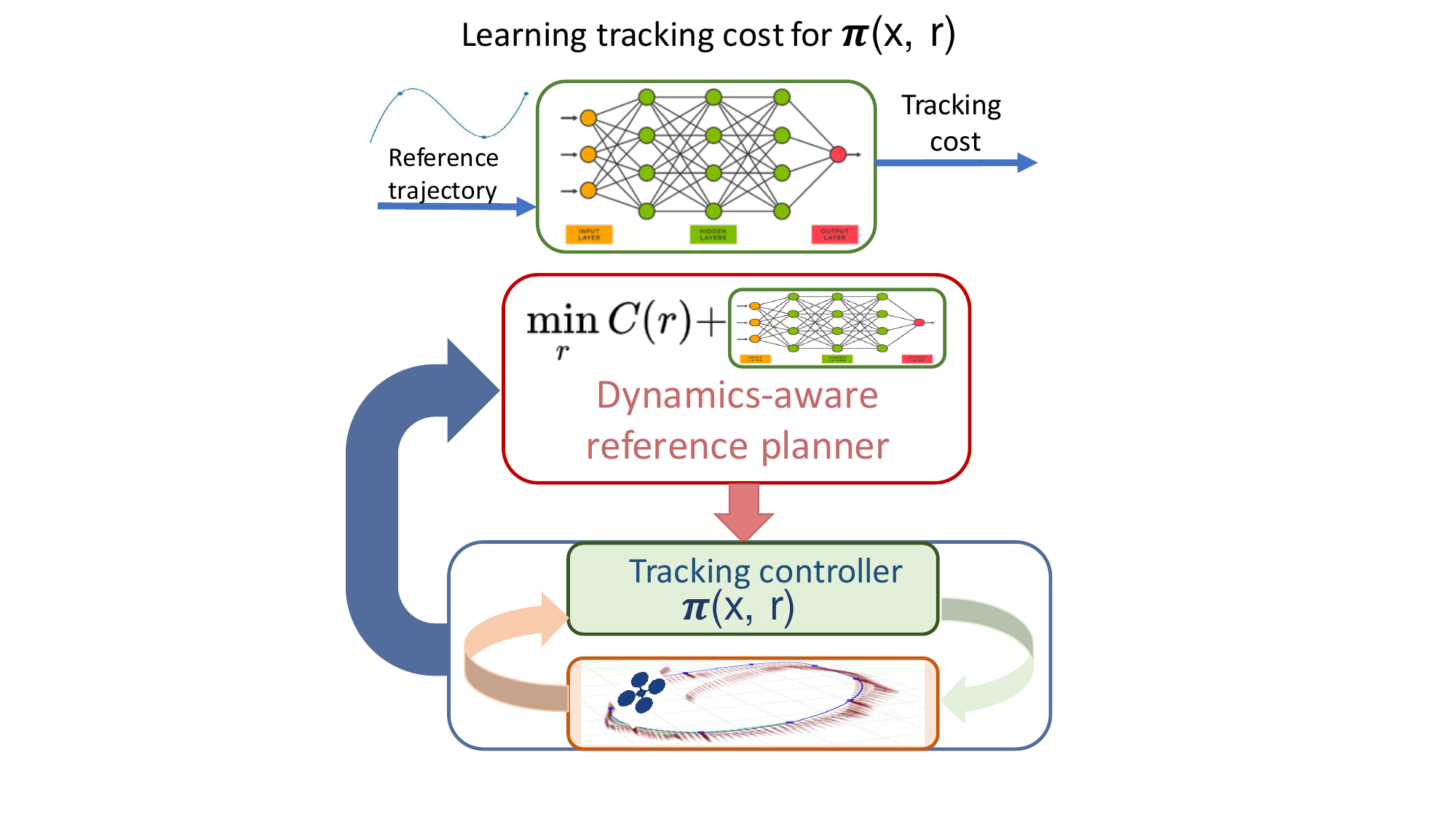}
    \caption{The figure shows our two-step framework. On top, we show the first step of learning the tracking cost function for any arbitrary policy $\pi(x, r)$. On the bottom, we show our \emph{dynamics-aware} planner that accounts for controller-based tracking cost planning reference trajectories.}
    \label{fig:layered-arch-quad}
\end{figure}

\subsection{Learning the tracking penalty through policy evaluation}
\label{sec:learning}
Computing the tracking penalty $g_\rho^{track}$ for general nonlinear dynamics and experimental hardware platforms with black-box feedback control policies is intractable. We therefore propose a supervised learning approach to learning the tracking penalty \eqref{eq:tracking-cost} from data as shown in Figure \ref{fig:layered-arch-quad}. 

We define the following augmented dynamical system with states $\mu_t \in \R^{(N+1)n}$ and control inputs $u_t \in \R^{k}$. 
The state $\mu_t$ is constructed by concatenating the nominal state $x_t$ and the reference trajectory $r_{t:t+N}$ of length $N$ starting at time $t$, i.e., $\mu_t = (x_t, r_{t:t+N})\in \R^{(N+1)n}$. Letting $\mu_t^x := x_t$ and $\mu_t^r := r_{t:t+N}$, the augmented system dynamics can be written as $\mu_{t+1} = h(\mu_t, u_t)$, where
\begin{equation}\label{eq:aug_dyn}
    h(\mu_t, u_t) :=  \begin{bmatrix} f(\mu_t^x, u_t) \\ Z \mu_t^r\end{bmatrix}.
\end{equation}
Here $Z \in \{0,1\}^{Nn \times Nn}$ is the block-upshift operator, i.e., a block matrix with $I_n$ along the first block super-diagonal, and zero elsewhere. The state $\mu_t^x=x_t$ evolves in exactly 
the same way as in the true dynamics~\eqref{eq:dynamics}, whereas the reference trajectory $\mu_t^r := r_{t:t+N}$ is shifted forward in time via $Z\mu_t^r= r_{t+1:t+1+N}$.  Fixing policy $\pi(\mu_t)$, we define the policy dependent tracking cost
\begin{equation}\label{eq:pi-tracking}
g_{\rho,\pi}^{track}(x_0, r_{0:N}) =
\sum_{t=0}^{N-1} \rho \left\lVert \mu_t^x - [\mu_t^r]_1 \right\rVert_2^2 + \lVert D_t u_t \rVert_2^2 + \rho \left\lVert \mu_N^x - [\mu_N^r]_1 \right\rVert_2^2.
\end{equation}
We note that this corresponds exactly to the objective function defining the tracking penalty~\eqref{eq:tracking-cost} evaluated under the control sequence $u_t = \pi(\mu_t).$  As such, the policy dependent tracking cost $g_{\rho,\pi}^{track}(x_0, r_{0:N})$ is naturally viewed as an upper-bound to the true optimal tracking cost, where the sub-optimality is dependent on the quality of the chosen policy $\pi$.  In particular, we have that $g_{\rho,\pi^\star}^{track} = g_{\rho}^{track}$ for any optimal policy $\pi^\star$ that solves the optimal control problem~\eqref{eq:tracking-cost}.


Noting that the policy dependent tracking penalty~\eqref{eq:pi-tracking} is defined in terms of stage-wise costs, we can interpret $g_{\rho,\pi}^{track}(x_0, r_{0:N})$ as a cost-to-go function associated with the Markov Decision Process defined by the cost~\eqref{eq:pi-tracking}, dynamics~\eqref{eq:aug_dyn}, and policy $\pi$.  We therefore use Monte Carlo sampling \cite{sutton2018reinforcement} to generate a set of $\mathcal{T}$ trajectories of horizon length $N$ given by $(x^{(i)}_{0:N}, u^{(i)}_{0:N-1}, r^{(i)}_{0:N})_{i=1}^{\mathcal{T}}$, 
where $x_{0:N}^{(i)}$ and $u^{(i)}_{0:N-1}$ are the $i$-th state and input trajectories collected from applying feedback control policy $\pi$ to track reference trajectories $r^{(i)}$.  We also compute the associated tracking cost labels $y^{(i)} := g_{\rho,\pi}^{track}(x_0^{(i)},r_{0:N}^{(i)})$.  We then used supervised learning to approximate the policy dependent tracking penalty~\eqref{eq:pi-tracking} by solving the following supervised learning problem 
$$
\begin{array}{rl}
     \mathrm{minimize}_{g\in\mathcal G} \ \sum_{i=1}^\mathcal{T}(g(x_0^{(i)},r_{0:N}^{(i)}) - y^{(i)})^2,
\end{array}
$$
over a suitable function class $\mathcal{G}$, e.g., feedforward neural networks, see \S\ref{sec:experiments} for more details.

\section{Dynamics-Aware Trajectory Generation for Under-Actuated \\Robotic Systems}
\label{sec:experiments}
We showed the flexibility of our framework by applying it to both a unicycle and a quadrotor control problem.  For each platform, we formulated a global planning and control problem, which is then subsequently relaxed according to the methods proposed in \S\ref{sec:layeredarch} to yield a dynamics-aware planning problem and a feedback control layer. We now evaluate our methods experimentally and demonstrate their effectiveness in simulation and in real-world experiments.

\subsection{Unicycle Control}
\label{sec:method-uni}


We consider the continuous time unicycle dynamics
$$\begin{bmatrix}
    \dot x_1 \\ \dot x_2 \\ \dot \theta
\end{bmatrix} = \left[\begin{array}{cc}
        \cos{\theta} & 0 \\
        \sin{\theta} & 0 \\
        0 & 1
    \end{array}\right] \left[\begin{array}{c}
         v \\
         \omega 
    \end{array}\right]
$$ where $(x_1,x_2)\in\R^2$ are the system's Cartesian coordinates,  $\theta$ is the heading angle, and $v$, $\omega$ are the instantaneous linear and angular velocities, respectively.  Letting $x = (x_1,x_2,\theta)$ and $u=(v, \omega)$, we can compactly write the dynamics as $\dot x = g^{cts}(x)u$, for suitably defined $g(x)\in\R^{3 \times 2}$. Letting $x_{t+1}=f_{uni}(x_t,u_t)$ be the \texttt{rk4} discretization of these continuous dynamics, we can then pose the global problem
\begin{equation}\label{eq:uni-global}
    \begin{array}{rl}
         \underset{{x_{0:N},u_{0:N-1}}}{\mathrm{minimize}}& \displaystyle\sum_{\tau\in\mathcal{T}_w} \|x_\tau - w_\tau\|_2^2 + \sum_{t=0}^{N-1} u_t^T R u_t  \\
         \text{subject to} &  x_{t+1}=f_{uni}(x_t,u_t),
    \end{array}
\end{equation}
where $w_\tau$ such that $\tau\in \mathcal{T}_w\subseteq \{0,\dots,N\}$ are waypoints that the unicycle should traverse at time $\tau$, and $R>0$ is a positive definite control cost matrix.

To instantiate the layering framework proposed in \S\ref{sec:layeredarch}, we fix a low layer continuous time feedback control policy as
$$
\pi_{uni}(x,r)=g(x)^{\dag}(\dot r + K_p(x-r)),
$$
We then define a new control input, $\Delta r=\dot r$, to obtain the continuous time closed-loop dynamics 
\begin{equation}
\begin{array}{rcl}
\dot x &=&  g(x)g(x)^\dag(\dot r + K_p(x-r))\\
\dot r &=& \Delta r
\end{array} . 
\end{equation}
Letting $\bar x := (x, r),$ we compactly rewrite the continuous time closed-loop dynamics as
\begin{equation}\label{eq:uni-cts-dyn}
\dot{\bar{x}}=f^{cts}_{\pi, uni}(\bar x, \Delta r)
\end{equation}
for an appropriately defined $f^{cts}_{\pi,uni}$.
Finally, we obtain the discrete time dynamics $f_{\pi,uni}(\bar x_t, \Delta r_t)$ used in the experiments below via a \texttt{rk4} discretization of the continuous time dynamics \eqref{eq:uni-cts-dyn}.

Given the fixed closed-loop dynamics using the policy $\pi_{uni}$, the dynamics-aware trajectory generation problem~\eqref{prob:layered-problem} is then given by
\begin{equation}\label{eq:uni-ref}
    \begin{array}{rl}
        \underset{r_{0:N}}{\mathrm{minimize}} &\displaystyle\sum_{\tau\in\mathcal{T}_w} \|r_\tau - w_\tau\|_2^2 + g_{\rho, \pi_{uni}}^{track}(x_0, r_{0:N}) 
    \end{array}
\end{equation}
where $g_{\rho, \pi_{uni}}^{track}(x_0, r_{0:N})$ is the policy dependent tracking penalty~\eqref{eq:pi-tracking} induced by the closed-loop dynamics $\bar x_{t+1} = f_{\pi,uni}(\bar x_t, \Delta r_t)$.

\paragraph{Data collection}
In order to estimate the policy dependent tracking penalty $g_{\rho, \pi_{uni}}^{track}(x_0, r_{0:N})$, we sample reference trajectories and roll them out on the closed-loop system.  In order to appropriately shape the landscape of the learned penalty, we sample both easy and difficult to track reference trajectories.  Towards that end, we generate \emph{easy to track} reference trajectories by using Iterative LQR (ilqr)~\cite{li2004iterative} to approximately solve the finite horizon constrained optimal control problem 
\begin{equation}\label{eq:uni-ocp}
    \begin{array}{rl}
 \underset{v_{0:N}, x_{0:N}}{\mathrm{minimize}} & \sum_{\tau\in\mathcal{T}_w} \|x_\tau - w_\tau\|_2^2 +\sum_{t=0}^{N-1} \Delta r_t^T R_w \Delta r_t  \\
        \text{s.t.} & \bar x_{t+1} = f_{\pi,uni}(\bar x_t, \Delta r_t)
        \\
        &x_0 = r_0,\ x_N = r_N 
    \end{array}
\end{equation}
where $R_w$ is a positive definite matrix penalizing variations in the reference trajectory. Additionally, we also generate state independent polynomial reference trajectories that are oblivious to the low layer closed-loop dynamics of the system and only satisfy the initial and terminal state constraints. This strikes a balance between having low cost but hard to compute ilqr trajectories and high cost but easy to compute polynomial trajectories. At inference, we solve a constrained optimization by applying gradient descent on the dynamics-aware trajectory generation problem~\eqref{eq:uni-ref}. 

We generate $500$ trajectories by sampling initial and goal locations from a uniform distribution over $[0, 2]^{2}$ and $[1, 3]^{2}$, respectively. Between each initial location and goal, we sample one waypoint by choosing a convex combination of the two points. The heading angles for the initial state is sampled at random from a uniform distribution on the interval $[0, \pi]$ and the goal heading angles are set to 0. As described in Section \ref{sec:method-uni}, we run the constrained ilqr algorithm on the closed loop dynamics \eqref{eq:uni-ocp} until convergence enforcing the initial and terminal state constraints. Additionally, we augment the training dataset with $500$ polynomial reference trajectories with randomly sampled initial, waypoint and goal conditions from the fixed intervals mentioned above. In this way, we include both easy to track trajectories (generated by ilqr) and difficult to track trajectories (polynomial) in order to appropriately shape the optimization landscape of the learned tracking penalty.  For testing, we generate $50$ trajectories with one or two waypoints each from the fixed intervals using the polynomial reference generation method described above.

\paragraph{Training} We train a multi-layer perceptron network with $3$ hidden layers of $\{1000, 500, 200\}$ neurons, respectively, with Exponential Linear Unit (ELU) activation functions. ---see~\nameref{sec:appendix} for more details. 
We train a separate network for each value of $\rho$ using a batch size of $64$, learning rate $10^{-4}$ and run for $2500$ epochs. The entire network is setup using the optimized \texttt{JAX} \cite{jax2018github}, \texttt{Optax} and \texttt{Flax} libraries. The loss function is optimized using stochastic gradient descent (SGD) with momentum set to $0.9$. At test time, i.e., when we compute trajectories to be tracked by the low-level controller, we freeze the weights of the network and run projected gradient descent (PGD) using \texttt{jaxopt} to locally solve the dynamics-aware trajectory planning problem~\eqref{eq:uni-ref}. We set the maximum number of iterations for projected gradient descent to $50$. 

\begin{figure}[t]
    \centering
    \includegraphics[width=\columnwidth]{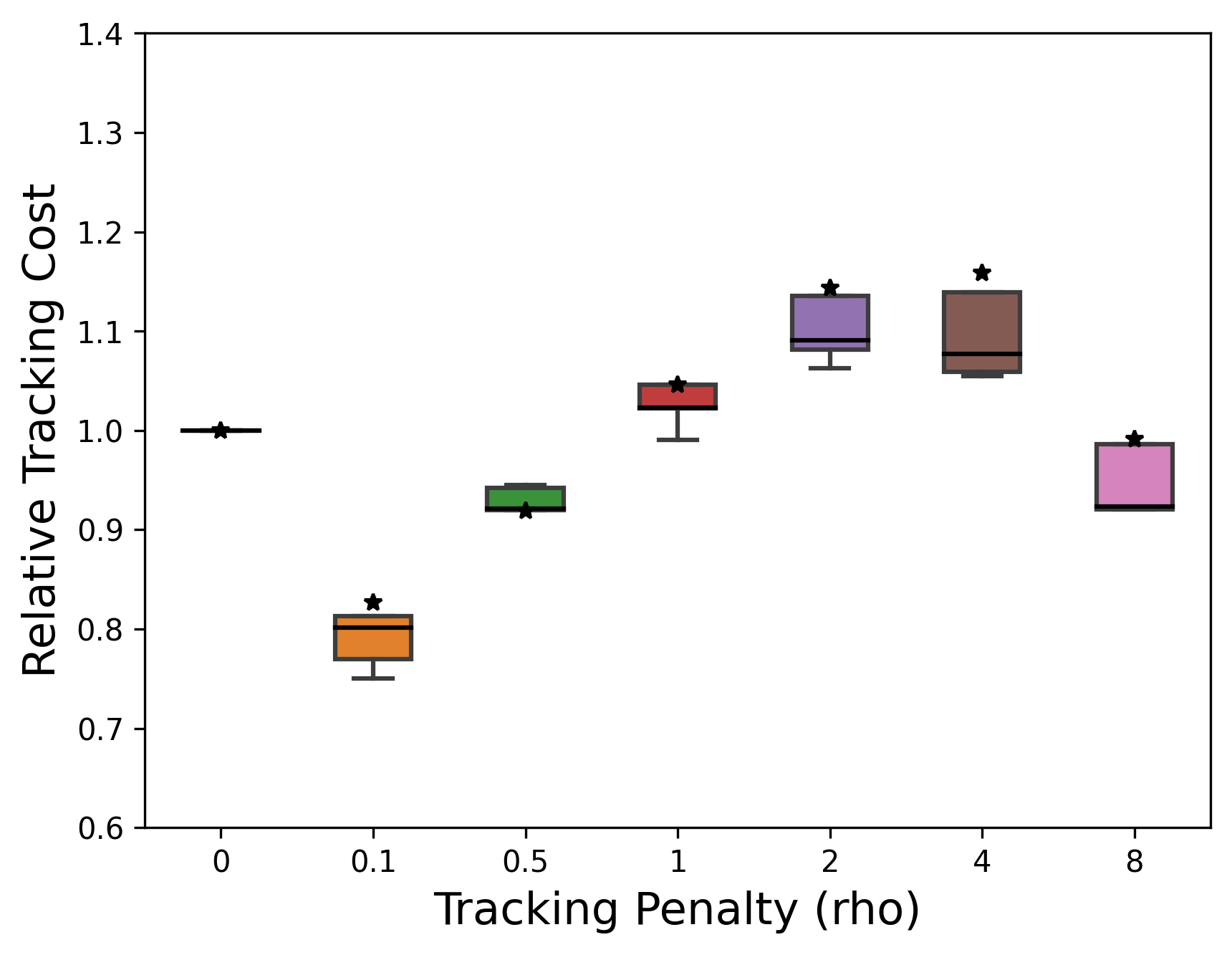}
    \caption{We show simulation results of the relative tracking cost on 50 trajectories for different values of $\rho$ (lower is better). $*$ represents the mean ratio of the dynamics-aware tracking cost to the polynomial reference tracking cost, the line dividing the boxes represent the median ratio separating the lower quartile from the upper quartile. The whiskers are the extreme values. We note that the tracking costs are obtained from the dynamics simulation (true cost) and not the network predictions.}
    \label{fig:cost-uni}
\end{figure}

\begin{figure}[t]
    \centering
    \includegraphics[width=\columnwidth]{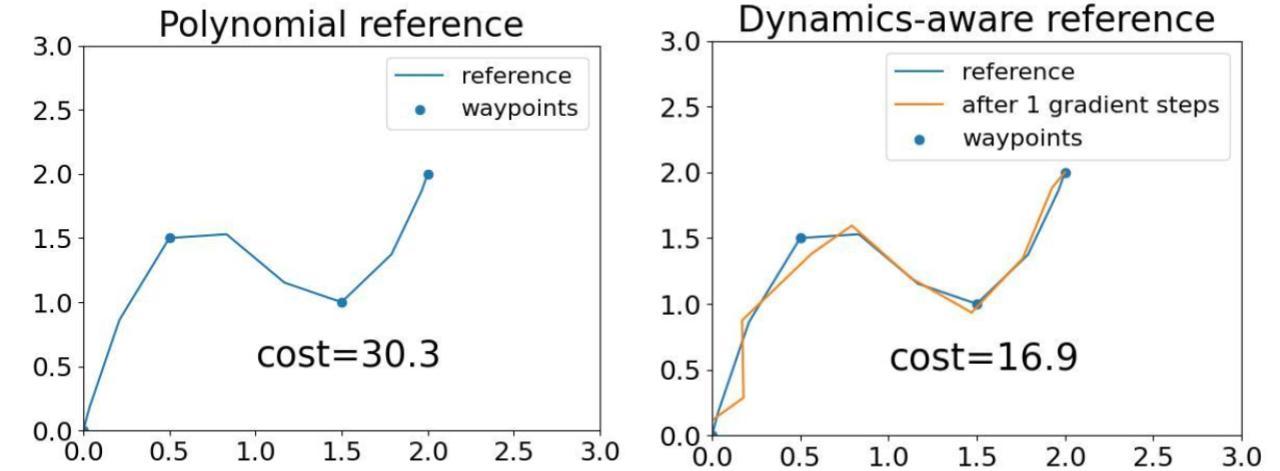}
    \caption{On the left, we show the initial polynomial reference trajectory in blue that is easy to compute but hard to track. On the right, we show the dynamics-aware trajectory in orange after one gradient step and demonstrate the tracking cost reduction from using our dynamics-aware planner.}
    \label{fig:pgd-uni}
\end{figure}

\paragraph{Results} We provide two types of evaluations on the learned policy dependent tracking penalty. First, we evaluate the network predictions for different values of relaxation weight $\rho$ on a test dataset consisting of $50$ trajectories generated independently and in an identical way to the training dataset. Next, we plot the relative tracking cost in Figure~\ref{fig:cost-uni}, where we compute the ratio of the tracking errors incurred by the trajectories returned by the dynamics-aware problem~\eqref{eq:uni-ref} to those incurred by polynomial interpolating (i.e., not dynamics-aware) trajectories---we emphasize these tracking costs are computed via rollouts of the actual closed-loop system on the trajectories. The lower the value of the relative cost, the more significant the tracking performance gain obtained from using our approach. Our results indicate that for appropriately chosen tracking weight $\rho>0$ the trajectories generated using our method are on average easier to track than polynomial interpolating trajectories. In Figure \ref{fig:pgd-uni}, we show the performance of our network ($\rho=0.1$) in planning reference trajectories that are of lower tracking cost (from the closed-loop dynamics simulation) with each gradient step. We also evaluated the average run time of our approach on $200$ trajectories and found that our algorithm is almost twice as fast as compared to the run time of the ilqr algorithm. Our network on average takes $6.9 \pm 0.7$ seconds to converge compared to ilqr which takes $11.4 \pm 0.5$ seconds. We conjecture that the results can be further improved by using convex parameterizations for the tracking penalty, such as input-convex-neural-networks (ICNN): we leave exploring this direction to future work. We also note that the approach is sensitive to the number of gradient steps based on the choice of $\rho$. 


\subsection{Quadrotor Control}
\label{sec:method-quad}
We consider the waypoint following problem where one is given a sequence of times $(\tau_i)_{i = 0}^W \in [0, T] \cap \mathbb{Z}$ and waypoints $\mathcal{W} = \{(p
_i, \psi_i)\}_{i = 0}^{W} \subseteq \mathbb{R}^3 \times S^1$, each specifying the desired position and yaw angle of the quadrotor. The goal is to generate a trajectory that passes through the waypoints at corresponding times. These conditions can be formulated in the OCP problem \eqref{prob:master-problem} by encoding the state constraints
\begin{equation*}
    p_{\tau_i} = p_i,\ \psi_{\tau_i} = \psi_i, \, \text{for } {i=0,\dots,W} 
\end{equation*}
within the constraint set $\mathcal{R}$. We consider a discrete time dynamical system $x_{t+1}=f_{q}(x_t,u_t)$  obtained by numerically integrating (using the Runge-Kutta scheme) a quadrotor with the following equations of motion:
\begin{equation}   \label{eq:quad_dynamics}
\dot x := 
\begin{bmatrix}
\dot{p} \\
\dot{v} \\
\dot{R} \\ 
\end{bmatrix}
=
\begin{bmatrix}
v \\ 
Re_3 c + g \\ 
R [\omega \times] \\ 
\end{bmatrix}.
\end{equation}
Here, the state $x$ of the agent consists of its position ($p \in \mathbb{R}^3$), velocity ($v \in \mathbb{R}^3$), and orientation ($R \in SO(3)$) with respect to the world frame, while the control input $u=(c,\omega)$ consist of total thrust, $c \in \mathbb{R}$, and angular velocity, $\omega \in \mathbb{R}^3$.

We can then pose the global problem that seeks to find a dynamically feasible \emph{minimium jerk} trajectory that passes through all of the waypoints:
\begin{equation}\label{eq:quad-global}
    \begin{array}{rl}
         \underset{{x_{0:N},u_{0:N-1}}}{\mathrm{minimize}}& \sum_{t=0}^{N-1} \|\dddot p_t\|_2^2 + \| \dot \psi_t\|_2^2 + \| u_t\|_2^2  \\
         \text{subject to} &  x_{t+1}=f_{q}(x_t,u_t), \\
         & p_{\tau_i} = p_i,\ \psi_{\tau_i} = \psi_i, \, \text{for } {i=0,\dots,W}
    \end{array}
\end{equation}

To instantiate the layering framework proposed in \S\ref{sec:layeredarch}, we fix a tracking control policy $\pi_{q}(x_t,r_{t:t+N})$ for the given dynamics $f_{q}(x_t,u_t)$, and a way to collect data on this tracking policy. In the experiments that follow, we use an SE(3) geometric controller \cite{SE3Controller}, but any tracking controller, e.g., a PID \cite{PIDController}, or even an RL-based controller~\cite{RLController}, can be equally accommodated by our framework.    Fixing the policy $\pi_q$, we can define the resulting closed-loop dynamics $x_{t+1}=f_{\pi,q}(x_t,r_{t:t+N})$ and corresponding policy dependent tracking penalty $g^{track}_{\rho,\pi_{q}}(x_0,r_{0:N})$ as in~\eqref{eq:pi-tracking}.

We now describe how to use the penalty $g^{track}_{\rho,\pi_{q}}(x_0,r_{0:N})$ to generate dynamics-aware trajectories that interpolate the waypoints $(p_i,\psi_i)$. First, we note that from differential flatness~\cite{fliess1995flatness,greeff2018flatness}, it suffices for us to generate trajectories for $x, y, z$, and $\psi$. We take the widely-adopted approach of parameterizing trajectories as piecewise polynomials of order $k_r$ that smoothly interpolates between waypoints. Specifically, each segment is parametrized by a polynomial
where $c_{i,k}^j$ denotes the $k$-th coefficient of polynomial $i$ for the dimension $j \in \{x, y, z, \psi\}$. 

With this parameterization, we can now recast the dynamics-aware trajectory generation problem~\eqref{prob:layered-problem} as one in the coefficients of the polynomial:
\begin{equation} \label{prob:quad-problem}
    \begin{array}{rl}
        \underset{\{c_{i,k}^j\}}{\mathrm{minimize}} &\displaystyle\sum_{\tau\in\mathcal{T}_w} \|r_\tau - w_\tau\|_2^2 + g_{\rho, \pi_q}^{track}(x_0, r_{0:N}) 
    \end{array}
\end{equation}
where $r_{0:N}$ is a linear map of the polynomial coefficients.

\paragraph{Data Collection} We generate $125$ trajectories by sampling the $x, y, z, \psi$ amplitudes of the Lissajous curves from a uniform distribution on the intervals $[-0.65, 0.65], [-0.55, 0.55], [-0.55, 0.55], [-0.6\pi, 0.6\pi]$, respectively. We select $5$ equally spaced waypoints on the Lissajous curves, parametrized by the following equations:
\begin{equation}\label{eq:lissajous}
    \begin{aligned}
    x_n = A_x\left(1 - \cos{\frac{2\pi n}{T}}\right),\
    y_n = A_y \left(\sin{\frac{2 \pi n}{T}}\right) \\
    z_n = A_z \left(\sin{\frac{2 \pi n}{T}}\right),\
    \psi_n = A_{\psi} \left(\sin{\frac{2 \pi n}{T}} \right)
    \end{aligned}
\end{equation}
where the time period of each trajectory is $3s$ and each second is discretized into 100 time steps. The discretization interval is selected based on the frequency of the feedback layer $SE(3)$ tracking controller~\cite{SE3Controller}, and $N=300$ is our planning horizon. We generate piece-wise polynomial reference trajectories $r_t \in \mathbb{R}^4$ denoting $x, y, z$ positions and $\psi$, for each segment between waypoints by minimizing the sum of squares of jerk and yaw angular velocity. We forward simulate the tracking controller using a state-of-the-art real-time quadrotor physics simulator on the ROS platform \cite{mohta2018fast} to record the system rollouts. 

\paragraph{Training} We train a multi-layer perceptron composed of 3 hidden layers with $\{500, 400, 200\}$ neurons, respectively, and ELU activation functions. ---see~\nameref{sec:appendix} for more details. 
Similar to the unicycle setup, we train a separate neural network for each value of $\rho$ using a batch size of 64, learning rate $10^{-3}$, and run for $2000$ epochs. The network implementation uses \texttt{Optax}, \texttt{Flax}, and \texttt{JAX} libraries for optimization and the loss function used is SGD with momentum set to be $0.9$. At test time, we use the `L-BFGS-B' solver from \texttt{jaxopt} to locally solve the dynamics-aware trajectory planning problem~\eqref{prob:quad-problem} where the objective function represents a trade-off between satisfying waypoints and the tracking cost of the $SE(3)$ geometric controller. We do no additional training for the hardware experiments. 

\begin{figure}[t]
    \centering
    \includegraphics[width=.8\columnwidth]{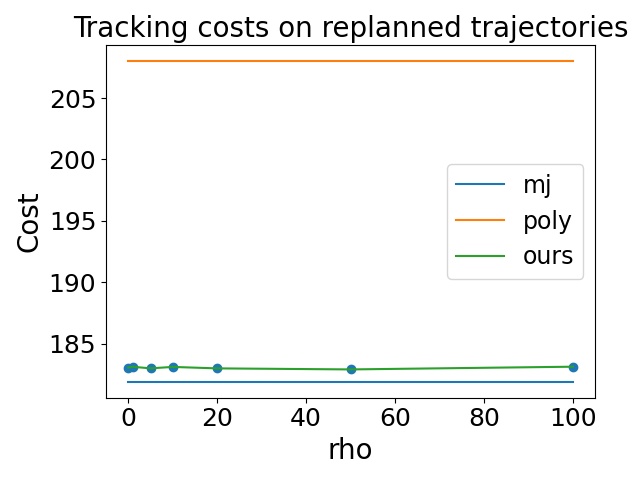}
    \caption{We show simulation results of the tracking cost on long-horizon trajectories of our approach in green against polynomial reference trajectories without smoothness constraints in orange for different values of $\rho$. The dots represent the mean of the costs. We demonstrate that our approach is able to achieve significantly lower tracking cost compared to the polynomial non-smooth trajectories.}
    \label{fig:cost-quad}
\end{figure}

\begin{figure}[t]
    \centering
    \includegraphics[width=\columnwidth]{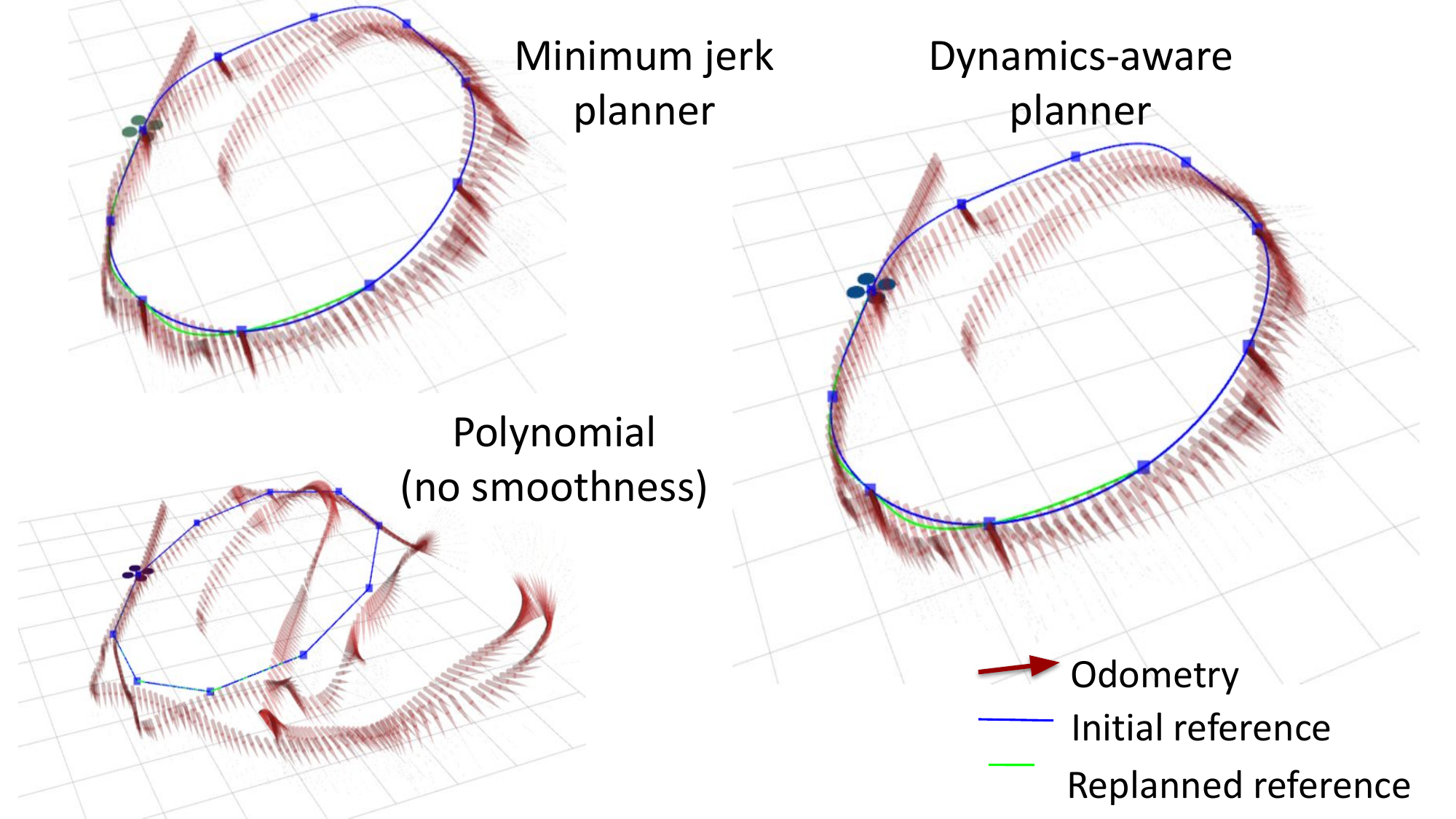}
    \caption{We show a comparison of the reference trajectory and the system rollouts using a state-of-the-art quadrotor physics simulator in the ROS platform. The blue lines and markers denote the planned reference and waypoints, in green we show the replanned short-horizon trajectories and the red arrows are the states from odometry.}
    \label{fig:traj-quad}
\end{figure}

\begin{figure}[t]
    \centering
    \includegraphics[width=\columnwidth]{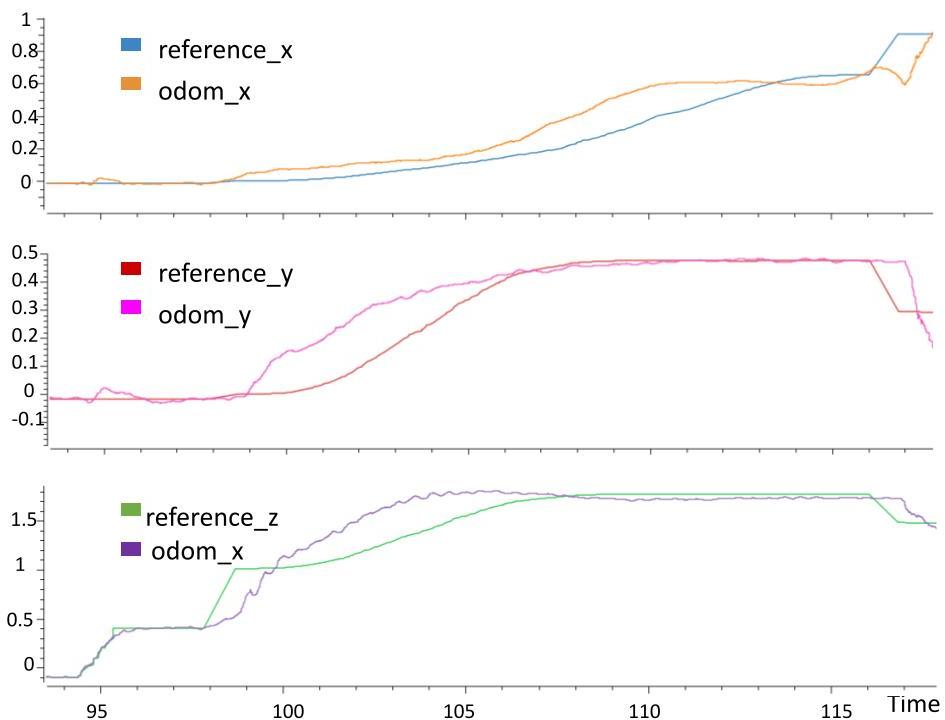}
    \caption{On running our dynamics-aware planner on the hardware platform, we plot the x, y, z reference curves and the odometry states for a 20 second trajectory. This demonstrates that our method handles the sim-to-real gap even without enforcing smoothness constraints at the planning layer.}
    \label{fig:traj-quad-hw}
\end{figure}

\paragraph{Results}
We evaluate the policy-dependent tracking penalty on trajectories generated independently and in an identical way to the training dataset, however, we allow for replanning after every $300$ time steps. We compare the tracking performance of our dynamics-aware framework with two trajectory generation methods, the standard minimum-jerk based planner satisfying constraints described in equations~\eqref{eq:quad-global} and polynomial trajectories that satisfy waypoint constraints but no smoothness constraints. Figure \ref{fig:traj-quad} shows the full path of the trajectory in blue, the replanned trajectories for every $300$ time steps in green and the red arrows correspond to the odometry states from the simulator. On the top left are results from using the minimum jerk planner, the bottom left shows the polynomial trajectories without smoothness constraints and on the right we show our dynamics-aware planner that solves the trajectory generation in~\eqref{prob:quad-problem}. Our planner is able to recover trajectories of low tracking cost by replanning with the learned tracking penalty every $300$ time steps. We also evaluate the tracking cost from the $SE(3)$ dynamics for different values of $\rho$, as shown in Figure \ref{fig:cost-quad}. We observe that the learned tracking penalty faithfully approximates the tracking cost function of the low layer $SE(3)$ feedback controller and that the dynamics-aware trajectories synthesized using the learned tracking penalty achieve a significant reduction in the tracking cost for every tracking weight value $\rho>0$ that we tested. Finally, as shown in Figure \ref{fig:traj-quad-hw}, we demonstrate our dynamics-aware trajectories on the Qualcomm-Snapdragon based hardware platform \cite{loianno2016estimation} to show that our method handles the sim-to-real gap without any additional training. 

\paragraph{Hardware} We use the hummingbird quadrotor platform running a VOXL Flight - PX4 Autonomy controller with on-board visual inertial odometry and inertial measurement unit (IMU) sensors for localization. We treat the quadrotor system as a remote work station and establish a communication interface using the ROS platform from a laptop to transmit the position commands for the low layer $SE(3)$ feedback controller. The position commands are $14$-dimensional vectors composed of position, velocity, acceleration, jerk, yaw angles and yaw angular speed computed using $x, y, z, \psi$ references. We generate trajectories online using our dynamics-aware planner and transmit the commands over WiFi to the quadrotor for execution and record the reference trajectory and the odometry states from each run. We plot the $x, y, z$ co-ordinates of the reference trajectories and odometry measurements across time as shown in Figure~\ref{fig:traj-quad-hw}. We note that our dynamics-aware framework is able to generate trajectories that are safe to be deployed and tracked by the $SE(3)$ controller even without enforcing smoothness constraints. In future work, we would like to eliminate latencies arising from communicating the commands over a network and aim towards running the dynamics-aware framework using the limited onboard compute of the quadrotor platform. 

\section{Conclusion} \label{sec:conclusion}
We showed that the familiar two layer architecture composed of a trajectory planning layer and a low-layer tracking controller can be derived via a suitable relaxation of a global optimization problem.  The result of this relaxation is a regularized trajectory planning problem, wherein the original state objective function is augmented with a tracking penalty which captures the low layer closed-loop system's ability to track a given reference trajectory.  We further observed that this penalty can be interpreted as the cost-to-go of an augmented system, and showed how it could be learned from data.  We demonstrated our results on waypoint tracking problems for a unicycle system and a quadrotor system in simulation and hardware. 
In both cases, our method yielded significantly easier to track trajectories than simple polynomial interpolations between waypoints.  Future work will look to develop more systematic approaches to collecting trajectory data for training the tracking penalty, and to derive statistical guarantees for the learned tracking penalty.

\newpage

\appendix
\section{Network Parameterization and Training}\label{sec:appendix}
We parameterize the tracking cost function as the exponential of a multi-layer perceptron $\phi(\cdot; \theta): \R^{(N+1)n} \to \R$ with parameter $\theta$ and minimize the loss between the labels and predictions 
by solving the empirical risk minimization problem
$\mathrm{minimize}_{g\in\mathcal G} \ \sum_{i=1}^\mathcal{T}(\phi(\mu^{(i)};\theta) - \log(y^{(i)}))^2$
where $\mathcal{T}$ is the number of collected trajectories, and $(\mu^{(i)}, y^{(i)})_{i=1}^\mathcal{T}$ are labeled pairs of the augmented state (containing the initial condition and reference trajectory) and the tracking cost it incurs (as described in \S\ref{sec:learning}).  At test time, we set the tracking penalty to be $\exp(\phi(\mu; \theta)$, thus ensuring that it is non-negative for all $\mu$. We found that this log reparameterization leads to more stable training of the network.  

\clearpage
\bibliographystyle{abbrvnat}

\bibliography{refs}

\begin{thebibliography}{35}
\providecommand{\natexlab}[1]{#1}
\providecommand{\url}[1]{\texttt{#1}}
\expandafter\ifx\csname urlstyle\endcsname\relax
  \providecommand{\doi}[1]{doi: #1}\else
  \providecommand{\doi}{doi: \begingroup \urlstyle{rm}\Url}\fi

\bibitem[Bradbury et~al.(2018)Bradbury, Frostig, Hawkins, Johnson, Leary,
  Maclaurin, and Wanderman-Milne]{jax2018github}
J.~Bradbury, R.~Frostig, P.~Hawkins, M.~J. Johnson, C.~Leary, D.~Maclaurin, and
  S.~Wanderman-Milne.
\newblock {JAX}: composable transformations of {P}ython+{N}um{P}y programs,
  2018.
\newblock URL \url{http://github.com/google/jax}.

\bibitem[Chiang et~al.(2007)Chiang, Low, Calderbank, and
  Doyle]{chiang2007layering}
M.~Chiang, S.~H. Low, A.~R. Calderbank, and J.~C. Doyle.
\newblock Layering as optimization decomposition: A mathematical theory of
  network architectures.
\newblock \emph{Proceedings of the IEEE}, 95\penalty0 (1):\penalty0 255--312,
  2007.

\bibitem[Drews et~al.(2017)Drews, Williams, Goldfain, Theodorou, and
  Rehg]{drews2017aggressive}
P.~Drews, G.~Williams, B.~Goldfain, E.~A. Theodorou, and J.~M. Rehg.
\newblock Aggressive deep driving: Combining convolutional neural networks and
  model predictive control.
\newblock In \emph{Conference on Robot Learning}, pages 133--142. PMLR, 2017.

\bibitem[Fliess et~al.(1995)Fliess, L{\'e}vine, Martin, and
  Rouchon]{fliess1995flatness}
M.~Fliess, J.~L{\'e}vine, P.~Martin, and P.~Rouchon.
\newblock Flatness and defect of non-linear systems: introductory theory and
  examples.
\newblock \emph{International journal of control}, 61\penalty0 (6):\penalty0
  1327--1361, 1995.

\bibitem[Greeff and Schoellig(2018)]{greeff2018flatness}
M.~Greeff and A.~P. Schoellig.
\newblock Flatness-based model predictive control for quadrotor trajectory
  tracking.
\newblock In \emph{2018 IEEE/RSJ International Conference on Intelligent Robots
  and Systems (IROS)}, pages 6740--6745. IEEE, 2018.

\bibitem[Gurriet et~al.(2018)Gurriet, Singletary, Reher, Ciarletta, Feron, and
  Ames]{gurriet2018towards}
T.~Gurriet, A.~Singletary, J.~Reher, L.~Ciarletta, E.~Feron, and A.~Ames.
\newblock Towards a framework for realizable safety critical control through
  active set invariance.
\newblock In \emph{2018 ACM/IEEE 9th International Conference on Cyber-Physical
  Systems (ICCPS)}, pages 98--106. IEEE, 2018.

\bibitem[Hehn and D’Andrea(2015)]{HehnDAndreaRealTimeTRO15}
M.~Hehn and R.~D’Andrea.
\newblock {R}eal-{T}ime {T}rajectory {G}eneration for {Q}uadrocopters.
\newblock \emph{IEEE Transactions on Robotics}, 31\penalty0 (4):\penalty0
  877--892, 2015.
\newblock \doi{10.1109/TRO.2015.2432611}.

\bibitem[Herbert et~al.(2017)Herbert, Chen, Han, Bansal, Fisac, and
  Tomlin]{herbert2017fastrack}
S.~L. Herbert, M.~Chen, S.~Han, S.~Bansal, J.~F. Fisac, and C.~J. Tomlin.
\newblock Fastrack: A modular framework for fast and guaranteed safe motion
  planning.
\newblock In \emph{2017 IEEE 56th Annual Conference on Decision and Control
  (CDC)}, pages 1517--1522. IEEE, 2017.

\bibitem[Kabzan et~al.(2019)Kabzan, Hewing, Liniger, and
  Zeilinger]{kabzan2019learning}
J.~Kabzan, L.~Hewing, A.~Liniger, and M.~N. Zeilinger.
\newblock Learning-based model predictive control for autonomous racing.
\newblock \emph{IEEE Robotics and Automation Letters}, 4\penalty0 (4):\penalty0
  3363--3370, 2019.

\bibitem[Kaufmann et~al.(2019)Kaufmann, Gehrig, Foehn, Ranftl, Dosovitskiy,
  Koltun, and Scaramuzza]{kaufmann2019beauty}
E.~Kaufmann, M.~Gehrig, P.~Foehn, R.~Ranftl, A.~Dosovitskiy, V.~Koltun, and
  D.~Scaramuzza.
\newblock Beauty and the beast: Optimal methods meet learning for drone racing.
\newblock In \emph{2019 International Conference on Robotics and Automation
  (ICRA)}, pages 690--696. IEEE, 2019.

\bibitem[Lambert et~al.(2019)Lambert, Drew, Yaconelli, Levine, Calandra, and
  Pister]{RLController}
N.~O. Lambert, D.~S. Drew, J.~Yaconelli, S.~Levine, R.~Calandra, and K.~S.~J.
  Pister.
\newblock Low-level control of a quadrotor with deep model-based reinforcement
  learning.
\newblock \emph{IEEE Robotics and Automation Letters}, 4\penalty0 (4):\penalty0
  4224--4230, 2019.
\newblock \doi{10.1109/LRA.2019.2930489}.

\bibitem[Lee et~al.(2010)Lee, Leok, and McClamroch]{SE3Controller}
T.~Lee, M.~Leok, and N.~H. McClamroch.
\newblock Geometric tracking control of a quadrotor uav on se(3).
\newblock In \emph{49th IEEE Conference on Decision and Control (CDC)}, pages
  5420--5425, 2010.
\newblock \doi{10.1109/CDC.2010.5717652}.

\bibitem[Levine and Abbeel(2014)]{levine2014learning}
S.~Levine and P.~Abbeel.
\newblock Learning neural network policies with guided policy search under
  unknown dynamics.
\newblock \emph{Advances in neural information processing systems}, 27, 2014.

\bibitem[Levine et~al.(2016)Levine, Finn, Darrell, and Abbeel]{levine2016end}
S.~Levine, C.~Finn, T.~Darrell, and P.~Abbeel.
\newblock End-to-end training of deep visuomotor policies.
\newblock \emph{The Journal of Machine Learning Research}, 17\penalty0
  (1):\penalty0 1334--1373, 2016.

\bibitem[Li and Todorov(2004)]{li2004iterative}
W.~Li and E.~Todorov.
\newblock Iterative linear quadratic regulator design for nonlinear biological
  movement systems.
\newblock In \emph{ICINCO (1)}, pages 222--229. Citeseer, 2004.

\bibitem[Liu et~al.(2018)Liu, Mohta, Atanasov, and Kumar]{LiuSE3SearchRAL18}
S.~Liu, K.~Mohta, N.~Atanasov, and V.~Kumar.
\newblock Search-{B}ased {M}otion {P}lanning for {A}ggressive {F}light in
  {S}{E}(3).
\newblock \emph{IEEE Robotics and Automation Letters}, 3\penalty0 (3):\penalty0
  2439--2446, 2018.
\newblock \doi{10.1109/LRA.2018.2795654}.

\bibitem[Loianno et~al.(2016)Loianno, Brunner, McGrath, and
  Kumar]{loianno2016estimation}
G.~Loianno, C.~Brunner, G.~McGrath, and V.~Kumar.
\newblock Estimation, control, and planning for aggressive flight with a small
  quadrotor with a single camera and imu.
\newblock \emph{IEEE Robotics and Automation Letters}, 2\penalty0 (2):\penalty0
  404--411, 2016.

\bibitem[Matni and Doyle(2016)]{matni2016theory}
N.~Matni and J.~C. Doyle.
\newblock A theory of dynamics, control and optimization in layered
  architectures.
\newblock In \emph{2016 American Control Conference (ACC)}, pages 2886--2893.
  IEEE, 2016.

\bibitem[Mellinger and Kumar(2011)]{mellinger2011minimum}
D.~Mellinger and V.~Kumar.
\newblock Minimum snap trajectory generation and control for quadrotors.
\newblock In \emph{2011 IEEE international conference on robotics and
  automation}, pages 2520--2525. IEEE, 2011.

\bibitem[Mohta et~al.(2018)Mohta, Watterson, Mulgaonkar, Liu, Qu, Makineni,
  Saulnier, Sun, Zhu, Delmerico, Thakur, Karydis, Atanasov, Loianno,
  Scaramuzza, Daniilidis, Taylor, and Kumar]{mohta2018fast}
K.~Mohta, M.~Watterson, Y.~Mulgaonkar, S.~Liu, C.~Qu, A.~Makineni, K.~Saulnier,
  K.~Sun, A.~Zhu, J.~Delmerico, D.~Thakur, K.~Karydis, N.~Atanasov, G.~Loianno,
  D.~Scaramuzza, K.~Daniilidis, C.~J. Taylor, and V.~Kumar.
\newblock Fast, autonomous flight in gps-denied and cluttered environments.
\newblock \emph{Journal of Field Robotics}, 35\penalty0 (1):\penalty0 101--120,
  2018.

\bibitem[Moreno-Valenzuela et~al.(2018)Moreno-Valenzuela, Pérez-Alcocer,
  Guerrero-Medina, and Dzul]{PIDController}
J.~Moreno-Valenzuela, R.~Pérez-Alcocer, M.~Guerrero-Medina, and A.~Dzul.
\newblock Nonlinear pid-type controller for quadrotor trajectory tracking.
\newblock \emph{IEEE/ASME Transactions on Mechatronics}, 23\penalty0
  (5):\penalty0 2436--2447, 2018.
\newblock \doi{10.1109/TMECH.2018.2855161}.

\bibitem[Mueller et~al.(2015)Mueller, Hehn, and
  D'Andrea]{MuellerDAndreaCompEffPrimitiveTRO15}
M.~W. Mueller, M.~Hehn, and R.~D'Andrea.
\newblock A {C}omputationally {E}fficient {M}otion {P}rimitive for
  {Q}uadrocopter {T}rajectory {G}eneration.
\newblock \emph{IEEE Transactions on Robotics}, 31\penalty0 (6):\penalty0
  1294--1310, 2015.
\newblock \doi{10.1109/TRO.2015.2479878}.

\bibitem[Ostafew et~al.(2016)Ostafew, Schoellig, and
  Barfoot]{ostafew2016robust}
C.~J. Ostafew, A.~P. Schoellig, and T.~D. Barfoot.
\newblock Robust constrained learning-based nmpc enabling reliable mobile robot
  path tracking.
\newblock \emph{The International Journal of Robotics Research}, 35\penalty0
  (13):\penalty0 1547--1563, 2016.

\bibitem[Paden et~al.(2016)Paden, {\v{C}}{\'a}p, Yong, Yershov, and
  Frazzoli]{paden2016survey}
B.~Paden, M.~{\v{C}}{\'a}p, S.~Z. Yong, D.~Yershov, and E.~Frazzoli.
\newblock A survey of motion planning and control techniques for self-driving
  urban vehicles.
\newblock \emph{IEEE Transactions on intelligent vehicles}, 1\penalty0
  (1):\penalty0 33--55, 2016.

\bibitem[Richter et~al.(2016)Richter, Bry, and Roy]{RichterBryRoy2016}
C.~Richter, A.~Bry, and N.~Roy.
\newblock \emph{Polynomial Trajectory Planning for Aggressive Quadrotor Flight
  in Dense Indoor Environments}, pages 649--666.
\newblock Springer International Publishing, Cham, 2016.
\newblock ISBN 978-3-319-28872-7.
\newblock \doi{10.1007/978-3-319-28872-7_37}.
\newblock URL \url{https://doi.org/10.1007/978-3-319-28872-7_37}.

\bibitem[Rosolia and Ames(2020)]{rosolia2020multi}
U.~Rosolia and A.~D. Ames.
\newblock Multi-rate control design leveraging control barrier functions and
  model predictive control policies.
\newblock \emph{IEEE Control Systems Letters}, 5\penalty0 (3):\penalty0
  1007--1012, 2020.

\bibitem[Rosolia and Borrelli(2019)]{rosolia2019learning}
U.~Rosolia and F.~Borrelli.
\newblock Learning how to autonomously race a car: a predictive control
  approach.
\newblock \emph{IEEE Transactions on Control Systems Technology}, 28\penalty0
  (6):\penalty0 2713--2719, 2019.

\bibitem[Roulet et~al.(2019)Roulet, Srinivasa, Drusvyatskiy, and
  Harchaoui]{roulet2019iterative}
V.~Roulet, S.~Srinivasa, D.~Drusvyatskiy, and Z.~Harchaoui.
\newblock Iterative linearized control: stable algorithms and complexity
  guarantees.
\newblock In \emph{International Conference on Machine Learning}, pages
  5518--5527. PMLR, 2019.

\bibitem[Singh et~al.(2017)Singh, Majumdar, Slotine, and
  Pavone]{singh2017robust}
S.~Singh, A.~Majumdar, J.-J. Slotine, and M.~Pavone.
\newblock Robust online motion planning via contraction theory and convex
  optimization.
\newblock In \emph{2017 IEEE International Conference on Robotics and
  Automation (ICRA)}, pages 5883--5890. IEEE, 2017.

\bibitem[Singh et~al.(2018)Singh, Chen, Herbert, Tomlin, and
  Pavone]{singh2018robust}
S.~Singh, M.~Chen, S.~L. Herbert, C.~J. Tomlin, and M.~Pavone.
\newblock Robust tracking with model mismatch for fast and safe planning: an
  sos optimization approach.
\newblock In \emph{International Workshop on the Algorithmic Foundations of
  Robotics}, pages 545--564. Springer, 2018.

\bibitem[Song and Scaramuzza(2022)]{song2022policy}
Y.~Song and D.~Scaramuzza.
\newblock Policy search for model predictive control with application to agile
  drone flight.
\newblock \emph{IEEE Transactions on Robotics}, 2022.

\bibitem[Sutton and Barto(2018)]{sutton2018reinforcement}
R.~S. Sutton and A.~G. Barto.
\newblock \emph{Reinforcement learning: An introduction}.
\newblock MIT press, 2018.

\bibitem[Wabersich and Zeilinger(2018)]{wabersich2018linear}
K.~P. Wabersich and M.~N. Zeilinger.
\newblock Linear model predictive safety certification for learning-based
  control.
\newblock In \emph{2018 IEEE Conference on Decision and Control (CDC)}, pages
  7130--7135. IEEE, 2018.

\bibitem[Williams et~al.(2017)Williams, Wagener, Goldfain, Drews, Rehg, Boots,
  and Theodorou]{williams2017information}
G.~Williams, N.~Wagener, B.~Goldfain, P.~Drews, J.~M. Rehg, B.~Boots, and E.~A.
  Theodorou.
\newblock Information theoretic mpc for model-based reinforcement learning.
\newblock In \emph{2017 IEEE International Conference on Robotics and
  Automation (ICRA)}, pages 1714--1721. IEEE, 2017.

\bibitem[Yin et~al.(2020)Yin, Bujarbaruah, Arcak, and
  Packard]{yin2020optimization}
H.~Yin, M.~Bujarbaruah, M.~Arcak, and A.~Packard.
\newblock Optimization based planner--tracker design for safety guarantees.
\newblock In \emph{2020 American Control Conference (ACC)}, pages 5194--5200.
  IEEE, 2020.

\end{thebibliography}

\end{document}